\documentclass[lettersize,journal]{IEEEtran}
\usepackage{amsmath,amsfonts}
\usepackage{algorithmic}
\usepackage{algorithm}
\usepackage{array}
\usepackage[caption=false,font=normalsize,labelfont=sf,textfont=sf]{subfig}
\usepackage{textcomp}
\usepackage{stfloats}
\usepackage{url}
\usepackage{balance}
\usepackage{verbatim}
\usepackage{graphicx}
\usepackage{cite}
\usepackage{color}
\usepackage{multirow}
\usepackage{booktabs}
\usepackage{tabularx}
\usepackage{ulem}
\usepackage{hyperref}
\usepackage{stfloats}
\usepackage{subfig}
\usepackage{rotating} 
\usepackage{ulem}  
\newcolumntype{C}{>{\centering\arraybackslash}X} 

\begin{document}

\title{Multi-Scale Cross-Fusion and Edge-Supervision Network for Image Splicing Localization}

\author{Yakun Niu, Pei Chen, Lei Zhang, Hongjian Yin, and Qi Chang~\IEEEmembership{}
\thanks{This work was supported in part by the National Natural Science Foundation of China (Grant 62202141 and Grant 62302135), in part by Key R\&D Projects in Henan Province (Grant 241111212800). (Corresponding author: Lei Zhang)}
\thanks{Yakun Niu, Pei Chen, Lei Zhang and Hongjian Yin are with the School of Computer and Information Engineering, Henan University, Kaifeng, 475000, China 
(e-mail: \{ykniu; chenpei; zhanglei; yinhongjian\}@henu.edu.cn).}
\thanks{Qi Chang is with the School of Cyberspace, Hangzhou Dianzi University, Hangzhou 310018, China (e-mail:qichang@hdu.edu.cn).}
}

\markboth{Journal of \LaTeX\ Class Files,~Vol.~14, No.~8, August~2021}%
{Shell \MakeLowercase{\textit{et al.}}: A Sample Article Using IEEEtran.cls for IEEE Journals}


\maketitle

\begin{abstract}

Image Splicing Localization (ISL) is a fundamental yet challenging task in digital forensics. Although current approaches have achieved promising performance, the edge information is insufficiently exploited, resulting in poor integrality and high false alarms. To tackle this problem, we propose a multi-scale cross-fusion and edge-supervision network for ISL.  Specifically, our framework consists of three key steps: multi-scale features cross-fusion, edge mask prediction and edge-supervision localization. Firstly, we input the RGB image and its noise image into a segmentation network to learn multi-scale features, which are then aggregated via a cross-scale fusion followed by a cross-domain fusion to enhance feature representation. Secondly, we design an edge mask prediction module to effectively mine the reliable boundary artifacts. Finally, the cross-fused features and the reliable edge mask information are seamlessly integrated via an attention mechanism to incrementally supervise and facilitate model training. Extensive experiments on publicly available datasets demonstrate that our proposed method is superior to state-of-the-art schemes.

\end{abstract}


\section{Introduction}
\IEEEPARstart{I}{n} recent years, with the wide application and popularization of image editing tools and multimedia technologies, the problem of digital image forgery has become more and more prominent. Image forgery can be divided into three categories: splicing{\cite{Splice, DNET, FARA}}, copy-move\cite{CopyMove, CopyMove1}, and removal\cite{Removal}, Among them, splicing represents the most prevalent method of image forgery. It involves the combination of a region of a donor image with a host image to create a natural and realistic forged image. Since image splicing forgery may propagate false or misleading information, it can seriously threaten our social security. Therefore, there is an urgent need to confirm the authenticity of an image and localize the forged regions. Due to the fact that, in spliced images, the forged and real regions come from different source images, the forgery can be detected by finding the statistic inconsistency between forged and real regions in an image \cite{Edge,TruFor}. 

Existing methods for ISL can be divided into traditional methods and deep learning-based methods. For traditional methods \cite{ADQ,Noise1,Noise2}, most of them rely on certain properties of the image and are only valid for images of a specific format, with poor generalization capabilities. For deep learning-based methods, with the application of Convolutional Neural Network\cite{CNN} (CNN) and Transformer\cite{Attention}, the accuracy of forgery localization has been significantly improved compared to traditional methods, especially in pixel-level localization. 
Wu \textit{et al.} proposed an end-to-end deep neural network, ManTra-Net\cite{ManTra}, by training a feature extractor with a self-supervised learning task. This enables it to detect multiple forms of forgery, including complex operations such as splicing and copy-move. 
Liu \textit{et al.}\cite{PSCC} proposed a new method aims to focus on utilizing spatial correlation and channel correlation in images to gradually extract features at different scales, thus helping to detect both simple and complex forgery. 
Guo \textit{et al.}\cite{HiFi} proposed a hierarchical fine-grained image forgery detection and localization method, which is able to cope with various means of forgery by capturing different types of image forgery traces through multi-level feature extraction and fine-grained image analysis. 
Xu \textit{et al.}\cite{FARA} proposed to perform feature aggregation at different levels of an image and use a region-aware learning mechanism to more accurately identify and locate the forged regions in an image. 
Liu \textit{et al.}\cite{DNET} proposed a dual encoder network (D-Net) to extract different types of features through two independent encoders and combines these features to identify the forged regions more accurately. 

However, most of the aforementioned image splicing forgery detection methods rely on specific types of features, such as RGB or noise features, which may not be able to fully capture the subtle forgery fingerprint. Moreover, the utilization of edge information is  insufficient, leading to missed or false detection.
To address these issues, we propose a multi-scale cross-fusion and edge-supervision network to capture subtle forgery fingerprint and learn rich edge information. First, we feed the RGB image and its noise image into the backbone network to learn multi-scale features, which are then aggregated through the cross-scale fusion and the cross-domain fusion to enhance the feature representation. Second, we extract edge artifact based on an edge mask prediction module from the learned multi-scale features. Finally, we use an attention mechanism to integrate the fused features and the edge mask to supervise splicing forgery localization.

\section{Proposed Method}
\label{Proposed Method}
\subsection{Overview}
\begin{figure}[t]
	\centering
	\includegraphics[width=3.5in]{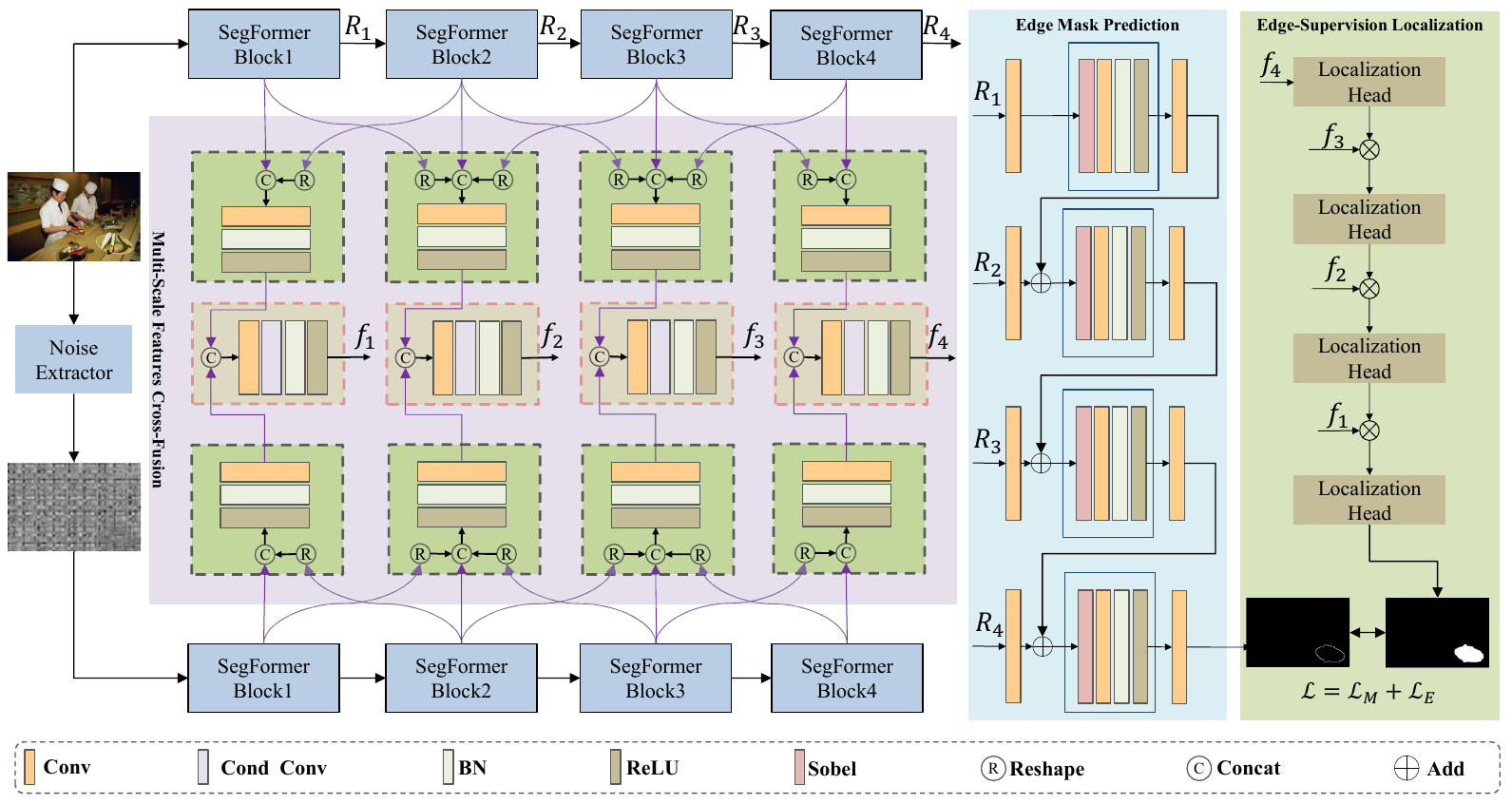}
	\caption{The proposed network architecture for splicing forgery localization. 
		}
	\label{model}
	\vspace{-1em}
\end{figure}

To fully capture subtle forgery fingerprint and edge information, we propose a novel dual-branch end-to-end network, as shown in Fig. \ref{model}. The network consists of three stages: multi-scale features cross-fusion, edge mask prediction, and edge-supervision localization. In the multi-scale features cross-fusion stage, for an input RGB image $X$, we first convert it into a noise image $X_N$ by NoisePrint++\cite{TruFor}. Then, both the RGB image and the noise image are fed into the backbone network for multi-scale features learning, respectively. The learned multi-scale features are further processed by the Cross-Scale Fusion (CSF) and the Cross-Domain Fusion (CDF) to enhance their representation. While CSF captures global and local features, CDF aims to exploit the complementarity between the RGB and noise domains. Then, we learn the reliable edge mask from the RGB domain features to fully mine the boundary artifacts for splicing forgery localization. In edge-supervision localization stage, the cross-fused features and the reliable edge mask are seamlessly integrated via an attention mechanism to incrementally supervise model training. 


\subsection{Multi-Scale Features Cross-Fusion}

Since forged areas have various sizes and irregular shapes, it is important to extract local and global features to handle the scale and shape variation for forgery localization. Therefore, we use SegFormer\cite{SegFormer} as the backbone of our proposed network. On the one hand, SegFormer is a powerful segmentation network, which is capable of efficiently handling multi-scale features in visual tasks. On the other hand, since image splicing forgery often involve subtle changes
in local regions, the ability of SegFormer to balance global and local features can significantly improve the localization performance, especially in complex splicing with abnormal regions.

Given an RGB image $X^{H\times W \times 3}$, we first input it into the backbone to obtain four different scale features respectively, denoted as $\{R_1^{H/2 \times W/2 \times 32},$ $R_2^{H/4 \times W/4 \times 64},$ $R_3^{H/8 \times W/8 \times 160},$ $ R_4^{H/16 \times W/16 \times 256}\}$. These four multi-scale RGB features are ones of their most intuitive visual features, focusing on macroscopic visual information such as color and texture of the image. As a result, they reflect different levels of information in the image. By acquiring the four multi-scale features, the network is able to capture different levels of information to form a complete representation of RGB features for more
comprehensive localization of splicing forgery. 


In addition, since the splicing forgery copies a region from an image and then pastes it to another image, the real and forged regions of a spliced image have different statistical characteristics, which can be also reflected in the noise space. Thus, we propose to use NoisePrint++ as the extractor to capture the noise fingerprint. The noise is a unique feature generated by the image sensor during the capture process, and it is very sensitive to forgery localization. Similar with RGB features,  the noise image $X_N$ is then fed into another SegFormer network to obtain multi-scale noise features at four different levels, denoted as $\left\{N_1,N_2,N_3,N_4\right\}$.

To learn hierarchical representations from the multi-scale features in both RGB and noise spaces, we propose two fusion strategies: cross-scale fusion and cross-domain fusion.

\subsubsection{Cross-Scale Fusion}
Features at different scales contain different levels of spatial information. That is, higher-resolution features can better capture the inconsistency between the boundaries of the forged region and the real region, while lower-resolution features help to effectively capture the global inconsistency between the forged region and the real region. By fusing neighborhood features, these detailed and global key information can be captured simultaneously, which helps to locate the forged region more accurately and enhances the model's ability to perceive both local and global information, thus improving the richness of feature representation.


Based on above analysis, we propose the CSF to aggregate the neighborhood of multi-scale features. We first resample the four multi-scale features to the same resolution and then concatenate them in the channel dimension. The number of channels is reduced by a $1\times 1$ convolution layer. The process of CSF for RGB features can be expressed as:
\begin{equation}  
f_{rgb}^i=\operatorname{ReLU}(\operatorname{BN}(\operatorname{Conv}(R_{i-1};R_{i};R_{i+1})))
\end{equation} 
where $f_{rgb}^i (i\in \{1,2,3,4\})$ denotes the fused RGB feature. If $i=1$, $R_1$ and $R_2$ are fused in CSF. Similarly, $f_{rgb}^4$ is obtained by fusing $R_3$ and $R_4$.

The splicing forged images often have obvious boundaries or unnatural transitions. By the fusion operation of convolution, the model's ability to perceive these boundary features can be enhanced, especially through channel compression and multi-scale fusion, which further refines the model's feature representation at the boundary and helps to more accurately distinguish the forged region from the original region.

It is worth noting that, similarly with RGB features, we also adopt CSF strategy to aggregate the multi-scale noise features to obtain the fused noise feature $f_{n}^i$. To save space, the process of fusion is not described in detail here.

\subsubsection{Cross-Domain Fusion}
RGB and noise features contain discriminative information: RGB features are good at describing the visible visual information of an image, whereas noise fingerprint features are more sensitive to the forged traces of an image. Therefore, these two different information are complementary in localizing the forged regions. With this in mind, we design the CDF based on conditional convolution (CondConv)\cite{CondConv} to fully mine the latent complementary relationships between RGB and noise features.


As shown in Fig. \ref{model}, we first concatenate them in the channel dimension, which can be viewed as a superposition of RGB and noise features. Then, the number of channels is reduced by  a convolutional layer with the convolutional kernel size of $1\times1$. With the aim of initially fusing the features of two different domains and reducing the computational complexity,
we then feed them into CondConv. The convolution kernel can be adaptively adjusted by CondConv according to the local features to amplify the feature differences and then more accurately differentiate the forged region from the real region. This dynamic convolution mechanism is exceptionally effective in revealing forged traces, as it possesses the capability to accurately capture inconsistence between real and forged regions, thereby enhancing the accuracy of localization. The process of CDF can be expressed as:
\begin{equation}
	f_i=\operatorname{ReLU}(\operatorname{BN}(\operatorname{CondConv}(\operatorname{Conv}(f_{rgb}^i;f_{n}^i))))
	\vspace{-0.5em}
\end{equation}

\subsection{Edge Mask Prediction}
In image splicing, even if the forger tries to make the spliced part blend well with the original image, there may still be inconsistencies in edge. Thus, the edge artifacts can be an important clues for forgery localization.

Inspired by\cite{MVSS}, we extract the edge artifact features of the forged region from the multi-scale features learned in the previous stage. The edge artifact features can make the network pay more attention to the forged part. For this purpose, we use the Sobel\cite{Sobel} edge operator and a convolutional layer with a 3$\times$3 kernel size to build an Edge Block (EB). As the prior information, the input of the current EB is the output of the previous one. Then it is multiplied by the RGB features of the corresponding scale to highlight the edge part with high weights. Finally, we input it into the next EB to gradually predict the final edge mask in a progressive way. This process can be expressed as:
\begin{equation}  
E_i=\operatorname{Conv}(\operatorname{EB}(\operatorname{Conv}(R_{i}\oplus E_{i-1})))
\end{equation} 
Here, for $i=1$, only $R_1$ is input to EB to have $E_1$.


\subsection{Edge-Supervision Localization}
In recent years, the attention mechanisms\cite{Attention} are widely used to locate the forged region. In this paper, we adopt the SCCM\cite{PSCC} as the localization head, which simultaneously incorporates spatial attention\cite{NonLocal} and channel attention\cite{SENet}. As shown in Fig. \ref{model}, the edge-supervision localization module contains a total of 4 SCCM heads. Similar to edge mask prediction stage, the predicted mask can be gradually generated in a progressive way: 
\begin{equation}  
M_{i}=\operatorname{SCCM}(\operatorname{reshape}(M_{i+1}) \otimes f_{i} )\\
\label{SCCM}
\end{equation} 
Here, $M_4 = \operatorname{SCCM}(f_{4})$ for $i=4$.

According to equation \ref{SCCM}, the current SCCM head uses the output of the previous one as the prior information to ultimately generate four predicted masks of different scales: $\left\{M_{1},M_{2},M_{3},M_{4}\right\}$. Here, the mask $M_{1}$ with the highest resolution is used as the final mask.

\subsection{Loss Function}
In our framework, the loss function consists of two parts: forgery mask loss and edge mask loss. The localization task can be regarded as a binary classification task, that is, to distinguish whether each pixel is a forged pixel or not. We use \textbf{0} to represent the pixels in the real region and \textbf{1} to represent the pixels in the forged region.

For the forgery mask, the binary cross-entropy loss $\left( \mathrm{BCE\_Loss} \right)$ is adopted for each localization head and they have the same weight. Since the edges of forged regions often only occupy a small part of an image, the sample distribution is extremely unbalanced. Therefore, the Dice loss is employed for the edge mask. Even if the number of edge pixels is small, dice loss can effectively measure the difference between the predicted edge and the real edge. Therefore, our final loss function is expressed as:
%
\begin{equation}
	L = \sum_{i=1}^{4} \text{BCE\_Loss}(M_i, G_i) + \text{Dice\_Loss}(E_{4}, G_{E})
\end{equation}
where $M_i$ and $G_i$ denote the predicted mask and the ground-truth in different scales respectively, $i \in \{1, 2, 3, 4\}$. $E_4$ and $G_{E}$ are the predicted edge-mask for $i=4$ and the corresponding ground-truth, respectively. Note that, we perform Sobel edge operator on $G_4$ to automatically obtain $G_E$.

\section{EXPERIMENTS}
\label{Experiments}
\subsection{Experimental setup}
\subsubsection{Datasets}
\label{dataset}
We first use images in DEFACTO\cite{DEFACTO} and PSCC\cite{PSCC}, a total of 120k images, to pre-train our model. Then, for fine-tune, we follow the same training/testing split on COLUMBIA\cite{Columbia}, CASIAv2\cite{CASIA} and NIST16\cite{NIST16} datasets, as in \cite{FARA} for fair comparisons.


\subsubsection{Experimental Setup}
Our model is implemented based on the PyTorch framework and experiments are performed using a NVIDIA GeForce RTX 3090 GPU. During the experiments, all training data were scaled to 256 $\times$ 256 pixels, the batch-size is set to 10, the learning rate is set to $2e-4$, and a total of 25 epochs of training are performed, with the learning rate halved every 5 epochs.

\subsubsection{Evaluation Metrics}
To quantify the localization performance,  $\mathrm { Precision }(P)$, $\mathrm { Recall }(R)$ and $\mathrm { F1 }$ scores are adopted as the evaluation metrics.
%
%

\subsection{Comparisons on Localization}
\label{Experiments Results}

To evaluate the performance of our proposed method, several state-of-the-art works: ManTra-Net\cite{ManTra}, PSCC-Net\cite{PSCC}, MVSS-Net++\cite{MVSS}, HiFi-Net\cite{HiFi}, FARA-Net\cite{FARA} and D-Net\cite{DNET} are selected as the baseline models for experimental comparisons. The results of all compared methods are either taken from their original papers or by running the publicly available source code.


As shown in Table \ref{Result}, our method achieves the best performance on the three datasets, and a large improvement in P, R and F1 is obtained compared with state-of-the-art methods. For example, when the Columbia dataset is employed, compared with recent method D-Net, our method improves Precision, Recall and F1 score by 0.5\%, 7.7\% and 4.1\% respectively. 
Besides, when the CASIAv2 is utilized, compared to the second best method PSCC-Net, our method achieves an improvement in the three metrics by 2.3\%, 4.2\% and 3.2\% respectively. Similarly, on NIST16, our method outperforms PSCC-Net by 2.1\%, 0.5\% and 2.3\%, respectively. Note that, the localization performance of all methods on CASIAv2 dataset is slight worse than those on the other two dataset. The reason is that the forged regions in the CASIAv2 dataset blend in well with their surrounding real regions. In our method, the edge information is fully employed, while the multi-scale features in both RGB and noise domains are deeply integrated, we can  effectively mine the inconsistency between forged regions and real regions in CASIAv2. 
\begin{table}[t]
	\centering
	\setlength{\tabcolsep}{3.3pt}
	\caption{Performance comparison of different methods.}
	\renewcommand{\arraystretch}{1.1}
	\begin{tabular}{lccccccccc}
		\toprule
		\multicolumn{1}{c}{\multirow{2}{*}{Methods}} & \multicolumn{3}{c}{CASIAv2} & \multicolumn{3}{c}{Columbia} & \multicolumn{3}{c}{NIST16}   \\
		\cmidrule(lr){2-4} \cmidrule(lr){5-7} \cmidrule(lr){8-10} 
		& P & R & F1  & P & R & F1 & P & R & F1 \\
		\midrule
		ManTra{\cite{ManTra}}& 82.1 & 79.3 & 80.7 & 85.6 & 84.9 & 85.2 & 81.6 & 82.4 & 82.0 \\
		PSCC-Net{\cite{PSCC}}& 87.6 & \uline{90.7} & \uline{89.1} & 83.1 & 79.7 & 81.4 & \uline{95.4} & 97.4 & \uline{96.4}  \\
		MVSS-Net++{\cite{MVSS}}& 86.4 & 85.1 & 85.7 & \uline{96.2} & 85.5 & 90.5 & 83.8 & 82.7 & 83.2 \\
		HiFi-Net{\cite{HiFi}}& 80.4 & 85.0 & 82.7 & 84.2 & 80.0 & 82.1 & 94.8 & \uline{99.4} & 96.2  \\
		FARA-Net{\cite{FARA}}& \uline{87.8} & 86.3 & 87.0 & 95.1 & 86.7 & 90.7 & 85.3 & 84.4 & 84.8 \\
		D-Net{\cite{DNET}}& 86.6 & 85.2 & 85.9 & 96.0 & \uline{90.1} & \uline{93.0} & 86.3 & 84.2 & 85.2 \\
		\midrule
		Ours & \textbf{89.9} & \textbf{94.9} & \textbf{92.3} & \textbf{96.5}  & \textbf{97.8} & \textbf{97.1} & \textbf{97.5} & \textbf{99.9} & \textbf{98.7} \\
		\bottomrule
	\end{tabular}
	\label{Result}
	\vspace{-0.5em}
\end{table}
\begin{figure}[t]
	\centering
	\includegraphics[width=3.2in]{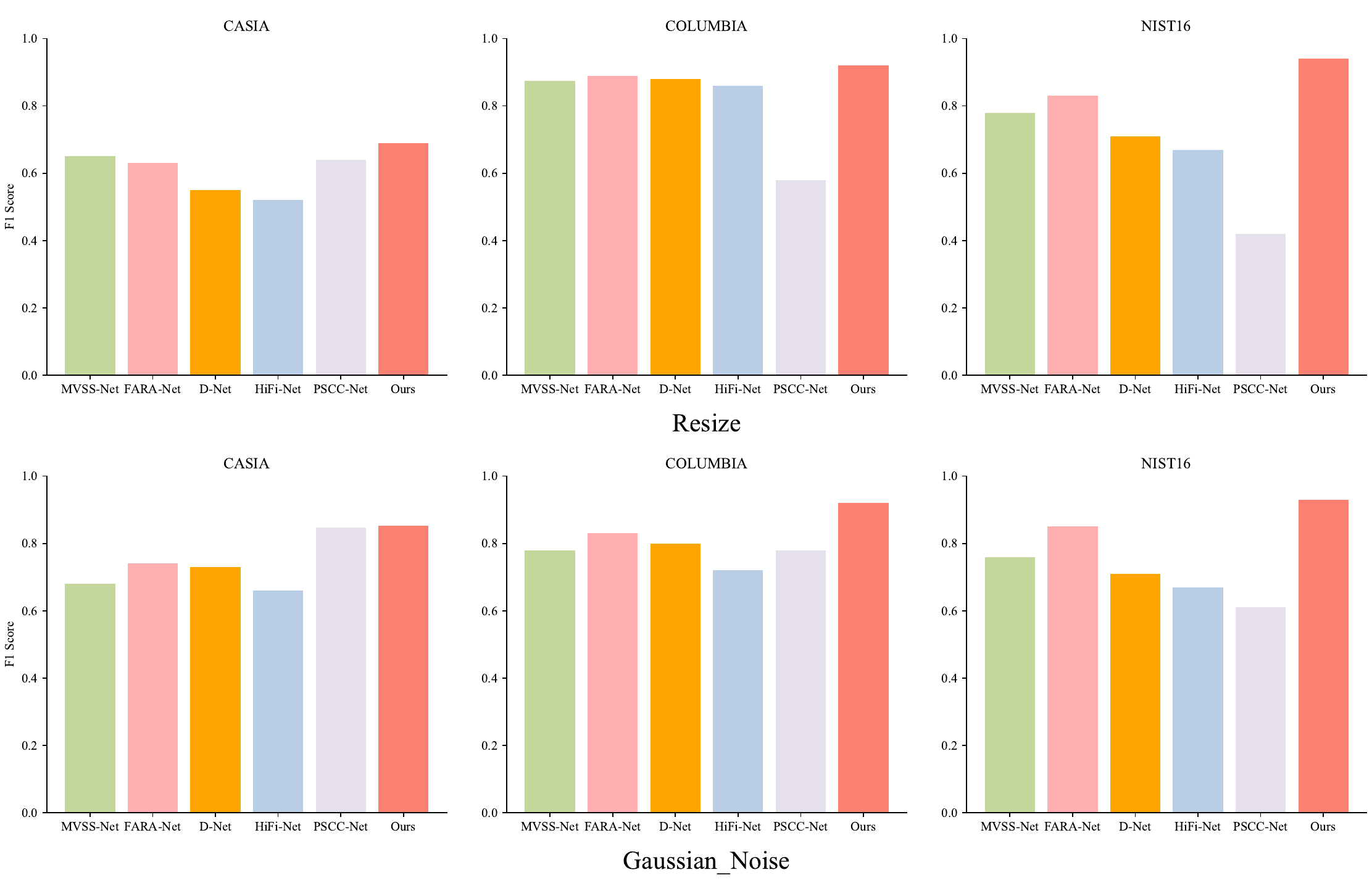}
	\caption{Analysis of robustness against image resize and Gaussian noise.}
	\label{rb}
	\vspace{-1em}
\end{figure}


In addition, we conducted a series of experiments to compare the robustness of our method with other methods. We attack the images using \textbf{Resize} with a ratio of 0.9 and \textbf{Gaussian Noise} with a variance of 3, respectively. It can be clearly seen from Fig. \ref{rb} that, our method achieves better performance on attacked images. This is due to the fact that our method makes full use of edge information and effectively integrates multi-scale features in both RGB and noise space.

\begin{figure}[t]
	\centering
	\includegraphics[width=3.2in]{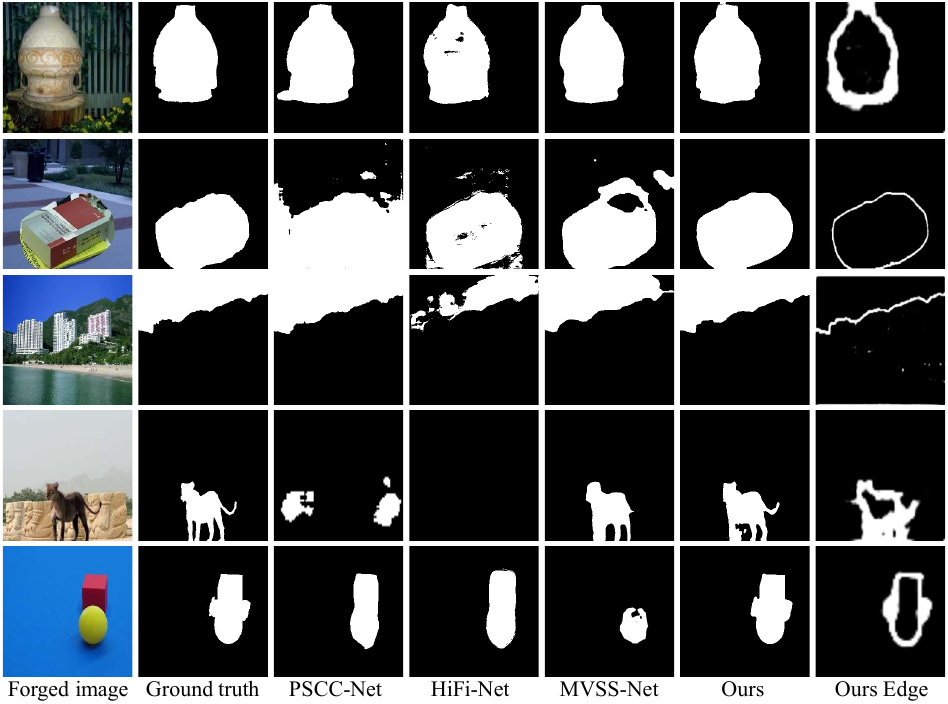}
	\caption{Examples of localization for different methods.}
	\label{result}
	\vspace{-1em}
\end{figure}

To further prove the effectiveness of our proposed method, some visualizations are given in Fig. \ref{result}. In our experiments, two aspects are visualized: the forgery prediction and the boundary prediction. It can be seen that our method can accurately detect the splicing forgery in both pixel-level and edge-level. Meaning that,  it has high recall and precision of forged regions. In addition, the proposed method is less sensitive to the scale variation. Both large (e.g., rows 1, 2) and small (e.g., rows 4, 5) forgery can be localized effectively.

\subsection{Ablation Study}
We performed a series of ablation experiments to study the contribution of each module to our proposed method, as shown in Table \ref{AB_Study}. 
We can see that our method incorporating different modules improve the performance, suggesting the critical of these modules in improving localization accuracy. Specifically, when the CSF is considered, the three metrics on the three datasets show an improvement of approximately 1.2\%. With the addition of CDF, the overall performance is improved by an average of about 0.5\%. Finally, after we additionally added the reliable predicted mask, the overall accuracy is improved by around 1.7\%. Therefore, training a network using edge information supervision can effectively make the network pay more attention to the characteristics of forged edges.

\begin{table}[!t]
	\centering
	\setlength{\tabcolsep}{2.8pt}
	\caption{Ablation Study Result
	}
	\renewcommand{\arraystretch}{1.1}
	\begin{tabular}{lccccccccc}
		\toprule
		\multicolumn{1}{c}{\multirow{2}{*}{Methods}} & \multicolumn{3}{c}{CASIAv2} & \multicolumn{3}{c}{Columbia} & \multicolumn{3}{c}{NIST16}   \\
		\cmidrule(lr){2-4} \cmidrule(lr){5-7} \cmidrule(lr){8-10}
		& P & R & F1  & P & R & F1 & P & R & F1  \\
		\midrule
		RGB      & 84.1 & 92.8 & 88.2 & 92.1 & 92.5 & 92.3 & 93.3 & 98.8 & 95.9 \\
		RGB\_NP & 85.8 & 92.4 & 88.9 & 92.2 & 93.7 & 93.0 & 94.5 & 97.8 & 96.1 \\
		RGB\_NP\_CSF  & \uline{89.0} & 92.5 & 90.7 & 92.8 & 94.1 & 93.4 & 95.6 & 99.4 & 97.5 \\
		RGB\_NP\_CSF\_CDF  & 88.8 & \uline{93.7} & \uline{91.2} & \uline{93.1} & \uline{94.9} & \uline{94.0} & \uline{96.4} & \uline{99.5} & \uline{97.9} \\
		\midrule
		Ours & \textbf{89.9} & \textbf{94.9} & \textbf{92.3} & \textbf{96.5}  & \textbf{97.8} & \textbf{97.1} & \textbf{97.5} & \textbf{99.9} & \textbf{98.7}\\
		\bottomrule
	\end{tabular}
	\label{AB_Study}
	\vspace{-0.5em}
\end{table}

\section{Conclusion}
\label{Conclusion}


In this paper, we proposed an end-to-end network for image splicing localization task by leveraging features cross-fusion as well as edge supervision. Our method consists of
three key steps: multi-scale features cross-fusion, edge mask prediction and edge-supervision localization. The multi-scale features cross-fusion stage successfully integrates multi-scale features from dual-domains. The pixel and noise inconsistencies between forged region and real region are employed in a mutually complementary way. Then, the reliable edge mask is effectively predicted to enhance the network's ability for fully capturing boundary artifacts. By an attention mechanism, the splicing forgery can be effectively localized by the way of edge-supervision. Experimental results validate the superiority of our method compared to state-of-the-art approaches.

\balance
\normalem
\bibliographystyle{unsrt}
\bibliography{ref}

\end{document}